\documentclass[letterpaper, 10 pt, journal, twoside]{IEEEtran}
\IEEEoverridecommandlockouts
\usepackage{amsmath,amsfonts}
\usepackage[caption=false,font=normalsize,labelfont=sf,textfont=sf]{subfig}
\usepackage{textcomp}
\usepackage[linesnumbered,ruled,vlined]{algorithm2e}

\usepackage{graphics}
\usepackage[pdftex]{graphicx}

\usepackage{epsfig}
\usepackage{mathptmx} 
\DeclareMathAlphabet{\mathcal}{OMS}{cmsy}{m}{n}
\DeclareSymbolFont{largesymbols}{OMX}{cmex}{m}{n}
\usepackage{times} 
\usepackage{amsmath}
\usepackage{amssymb}
\usepackage{float}  
\usepackage{amsmath,bm}
\usepackage{cite}
\usepackage{CJKutf8}
\usepackage{booktabs}
\usepackage{url}
\usepackage{bm}
\usepackage{multirow}
\usepackage{booktabs}
\usepackage{epstopdf}
\usepackage{epsfig}
\usepackage{longtable}
\usepackage{supertabular}
\usepackage{changepage}
\usepackage{verbatim}
\usepackage{xcolor}
\usepackage{subfig}
\usepackage{graphicx}
\usepackage{float}
\usepackage{stfloats}
\usepackage[mathscr]{euscript}
\usepackage[ruled]{algorithm2e}
\usepackage{array}
\usepackage{url}
\usepackage{tikz,xcolor}

\usepackage{times}
\usepackage[pdftex]{graphicx}
\usepackage{amsmath,amssymb,amsopn,amstext,amsfonts}
\usepackage{cancel}
\usepackage{balance}
\usepackage{color}
\usepackage{mathtools}
\usepackage[ruled,vlined]{algorithm2e}
\usepackage{bm}

\usepackage{diagbox}
\usepackage{float}
\usepackage{epstopdf}
\usepackage{pifont}
\usepackage{fixltx2e}
\usepackage{amsmath}
\usepackage{multirow}
\usepackage{booktabs}
\usepackage{url}
\usepackage{svg}
\usepackage{threeparttable}
\usepackage[linkcolor=black,citecolor=black,urlcolor=black,colorlinks=true]{hyperref}
\usepackage{subcaption}
\usepackage{slashbox}
\usepackage{makecell}

\usepackage{soul} 
\DeclareGraphicsExtensions{.png,.jpg,.eps,.pdf}

\title{An Efficient Trajectory Generation for Bi-copter Flight in Tight Space}

\author{Xin Dong, Yangjie Cui, Jingwu Xiang, Daochun Li$^{*}$, and Zhan Tu$^{*}$

\thanks{The authors are with the School of Aerospace Science and Engineering, Beihang University, Beijing 100191, China.}%
\thanks{Jingwu Xiang is also with the Tianmushan Laboratory Xixi Octagon City, Yuhang District, Hangzhou 310023, China.}%
\thanks{Zhan Tu is also with the Institute of Unmanned System, Beihang University, Beijing 100191, China. {\tt\footnotesize Email: zhantu@buaa.edu.cn}}%
\thanks{\textsuperscript{$*$} Author for correspondence.}
}

\UseRawInputEncoding
\let\oldtwocolumn\twocolumn
\renewcommand\twocolumn[1][]{%
    \oldtwocolumn[{#1}{
    \begin{center}
            \vspace{-0.8cm} 
           \includegraphics[width=.9\textwidth]{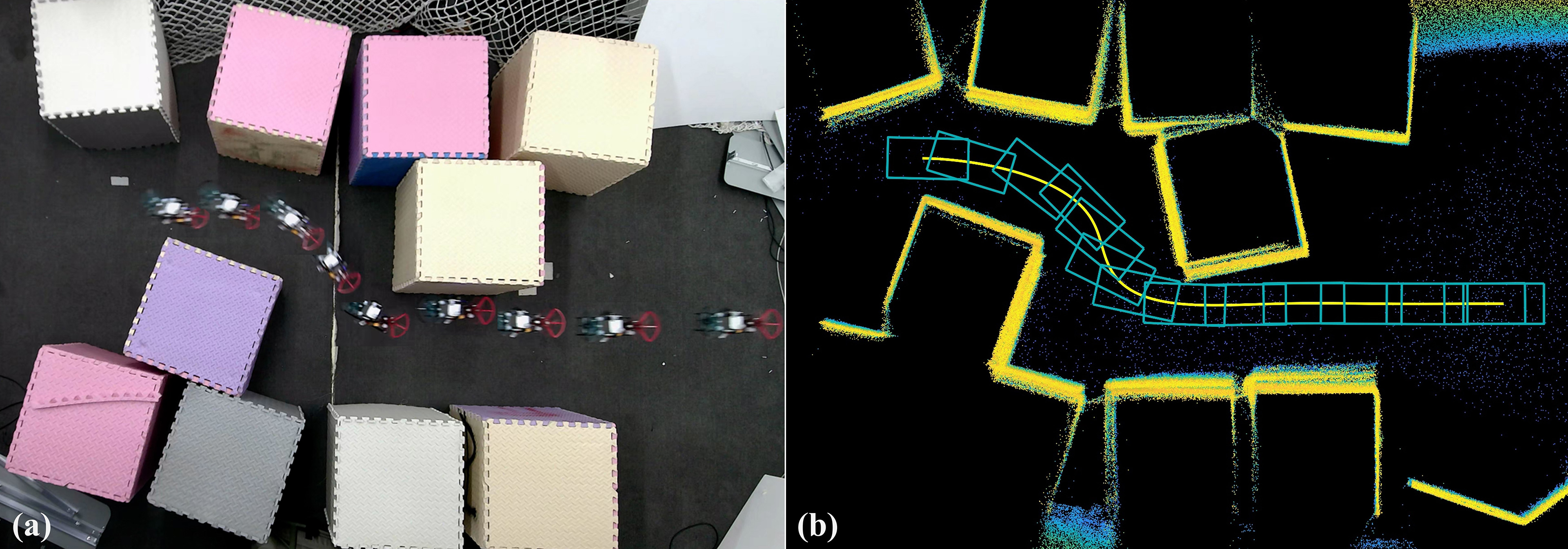}
           \captionof{figure}{\label{fig:top} Real-world experiment of bi-copter motion planning in narrow environments. (a) Real-world flight of bicopter. (b) Trajectory planning in point cloud map.}
           \label{fig:show_result}
        \end{center}
    }]
}
\begin{document}
 \makeatother
\maketitle
\setcounter{figure}{1}
 

\begin{abstract}

Unlike squared (or alike) quadrotors, elongated bi-copters leverage natural superiority in crossing tight spaces. To date, extensive works have focused on the design, modeling, and control of bi-copters. Besides, a proper motion planner utilizing bi-copters' shape characteristics is essential to efficiently and safely traverse tight spaces, yet it has rarely been studied. Current motion planning methods will significantly compromise their ability to traverse narrow spaces if the map is inflated based on the long dimension of the bi-copter. In this paper, we propose an efficient motion planning method that enables the safe navigation of bi-copters through narrow spaces. We first adapt a dynamic, feasible path-finding algorithm with whole-body collision checks to generate a collision-free path. Subsequently, we jointly optimize the position and rotation of the bi-copter to produce a trajectory that is safe, dynamically feasible, and smooth. Extensive simulations and real-world experiments have been conducted to verify the reliability and robustness of the proposed method. 
\end{abstract}


\IEEEpeerreviewmaketitle

\section{Introduction}

\IEEEPARstart{I}{n} recent years, due to their high agility and maneuverability, micro UAVs have demonstrated remarkable proficiency in rescue operations, exploration, indoor mapping, and many other fields \cite{mahony2012multirotor, zhou2021fuel,feng2023predrecon,li2023autotrans}. Quadcopters have been widely adopted among various micro UAV prototypes due to their mechanical simplicity and differential flatness properties \cite{Faessler18ral}. However, the traversability of quadrotors is severely limited by their four-propeller design, and aggressive maneuvers or active deformation must be considered to cross the narrow gaps.
To overcome this limitation, numerous strategies have been proposed for quadrotors. Han et al. \cite{han2021fastracing} enabled quadrotors to traverse narrow gaps by generating aggressive trajectories, allowing them to pass through at steep angles. However, such extreme attack angles can only be maintained briefly, rendering the approach impractical for long, narrow corridor environments. Alternatively, Fabris et al. \cite{fabris2021geometry} adapted a morphing mechanism to dynamically adjust the quadrotor's body size for narrow space navigation. Yet, the additional morphing mechanism may further compromise the reliability and payload capacity of such quadrotors.

 
To further extend the ability to navigate narrow spaces, a more compact prototype of multi-rotor UAVs, known as the bi-copter, has been proposed \cite{qin2020gemini,qin2022gemini2,he2022design}. Equipped with only two rotors driven by two motors, the bi-copter has demonstrated significant advantages in terms of load capacity and endurance \cite{qin2020gemini}. Crucially, with only two propellers, the width of the bi-copter can be limited to the diameter of the propellers, thereby enhancing its traversability in complex environments. To date, the related research of the bi-copter UAVs is mainly focused on design, modeling, and flight control. In contrast, the motion planning method of bi-copter UAVs that focuses on their unique traversability properties has rarely been studied systematically.

\begin{figure}[t]
     \centering
      \includegraphics[width=\linewidth]{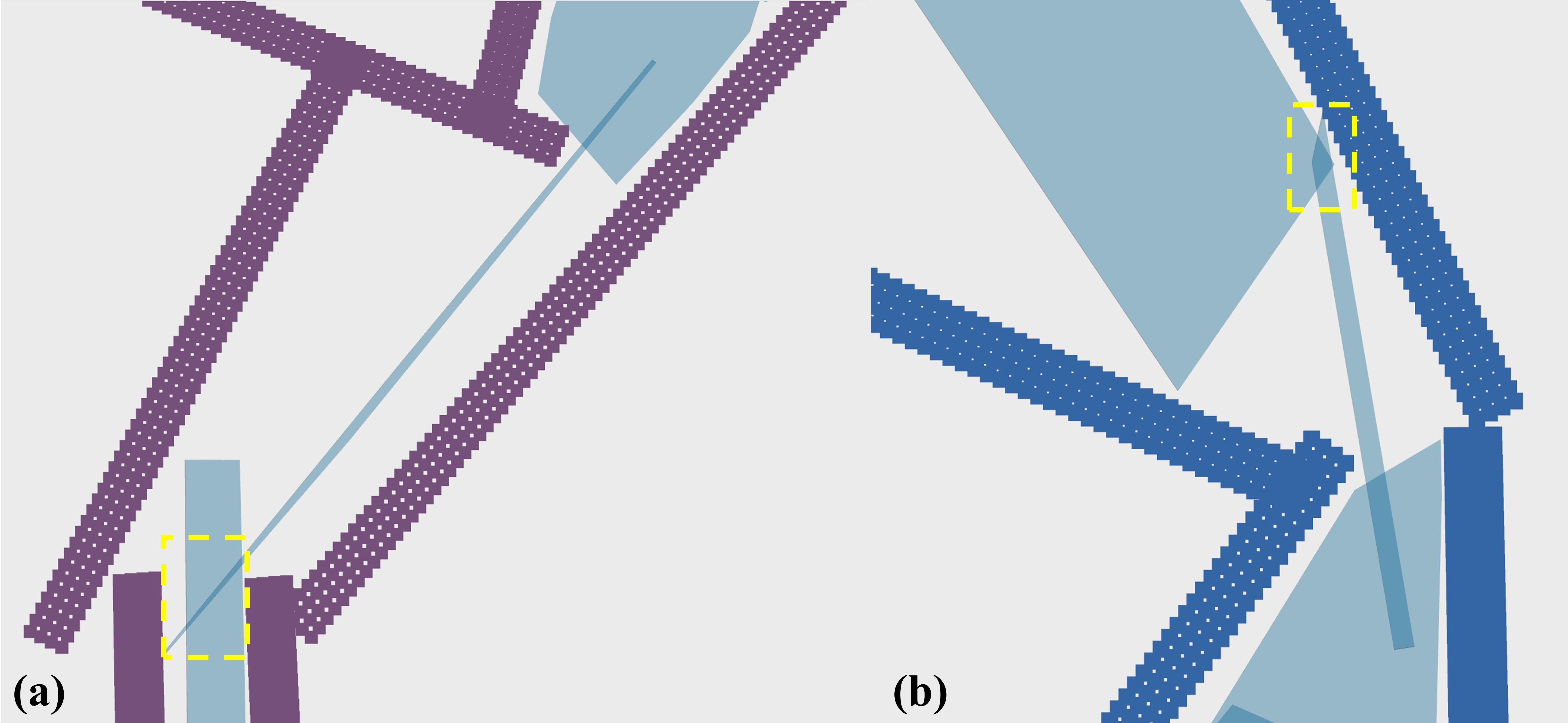}
      \caption{Flight corridor generation in narrow environments. (a): The intersection of two adjacent polyhedrons is too slim, failing to accommodate the bi-copter's body width. (b): The intersection of two adjacent polyhedrons is insufficiently small. }
      \label{fig: narrow_com}
\end{figure}

Traditional motion planning algorithms inflate the environmental grid map and model the UAV as a mass-point model \cite{zhou2021ego,tordesillas2019faster,penicka2022minimum,zhou2019fast,minisnap2011kumar}, which is suitable for conventional symmetrical multi-rotor UAVs in simple environments. However, treating them as mere mass points is inadequate for rectangular-shaped robots like bi-copter UAVs navigating complex, narrow spaces. Setting the inflation size to the longer dimension of the bi-copter eliminates many navigable gaps while setting it to the shorter dimension risks collisions in narrow environments. Therefore, the robot's shape must be factored into collision evaluation strategies to utilize the bi-copter's traversability fully. 
As for whole-body motion planning,  aside from strategies that adopt highly aggressive trajectories requiring UAVs to perform maneuvers with large flip angles \cite{han2021fastracing,loianno2017estimation}, approaches are primarily categorized into ESDF-based \cite{oehler2020whole,mittal2022articulated} and corridor-based methods \cite{li2022optimization,ji2022real,han2023efficient,ma2023decentralized}. ESDF-based methods need to maintain an Euclidean Signed Distance Field within a certain range, which can be computationally expensive. For corridor-based methods, the non-collision space is divided by a series of polyhedrons to form a flight corridor, constraining the robot from moving within these polyhedrons. However, in narrow environments, the generation of a flight corridor may fail. Moreover, the intersection of polyhedrons in these corridors often cannot contain an entire robot in such constricted spaces, as shown in Fig. \ref{fig: narrow_com}, presenting a significant challenge to their application in narrow environments.

In summary, while bi-copters show exceptional traversability in narrow environments, most of current research has focused on platform design, modeling, and control algorithms \cite{qin2020gemini,qin2022gemini2,he2022design,abedini2021robust,li2020modeling,albayrak2019design}. However, adapting mass-point-based motion planning algorithms demonstrates infeasible for narrow environments to rectangular platforms. To address this open challenge, we propose a whole-body bi-copter trajectory planning algorithm to achieve fast and robust collision-free trajectory planning in a complex, narrow corridor environment. Initially, we adopt a whole-body dynamic feasible path-finding algorithm to identify collision-free paths in narrow environments. Building upon this path, we jointly optimize the bi-copter's position and orientation regarding whole-body collision-free, trajectory smoothness, and dynamic feasibility. To verify the 
reliability and robustness of the proposed method, extensive experiments of both simulated and real-world were conducted. The main contributions of our papers are as follows:

\begin{itemize}
    \item We proposed a whole-body collision-free dynamic feasible path search algorithm for a rectangle-shaped bi-copter with planni position and yaw.
    \item Based on the initial path, we jointly optimize the position and yaw trajectory in narrow environments simultaneously to generate a smooth, dynamic, feasible, and collision-free trajectory.
    \item Extensive simulation and real-world experiments were applied to validate the reliability of the proposed method in different narrow environments.
\end{itemize}

The remainder of the paper is organized as follows:
Section \ref{sec:model} introduces the bi-copter platform and the differential flatness property of the bi-copter. Section \ref{sec:path} addresses the whole-body collision-free dynamic feasible path planning algorithm. 
Section \ref{sec:experiments} presents the simulation and real-world experiments based ons the proposed method. Finally, Section \ref{sec:conclusion} concludes the remarks of our work.

\section{System Overview}
\label{sec:model}
\subsection{Test Bi-copter Vehicle}

\begin{figure}[t]
     \centering
      \includegraphics[width=\linewidth]{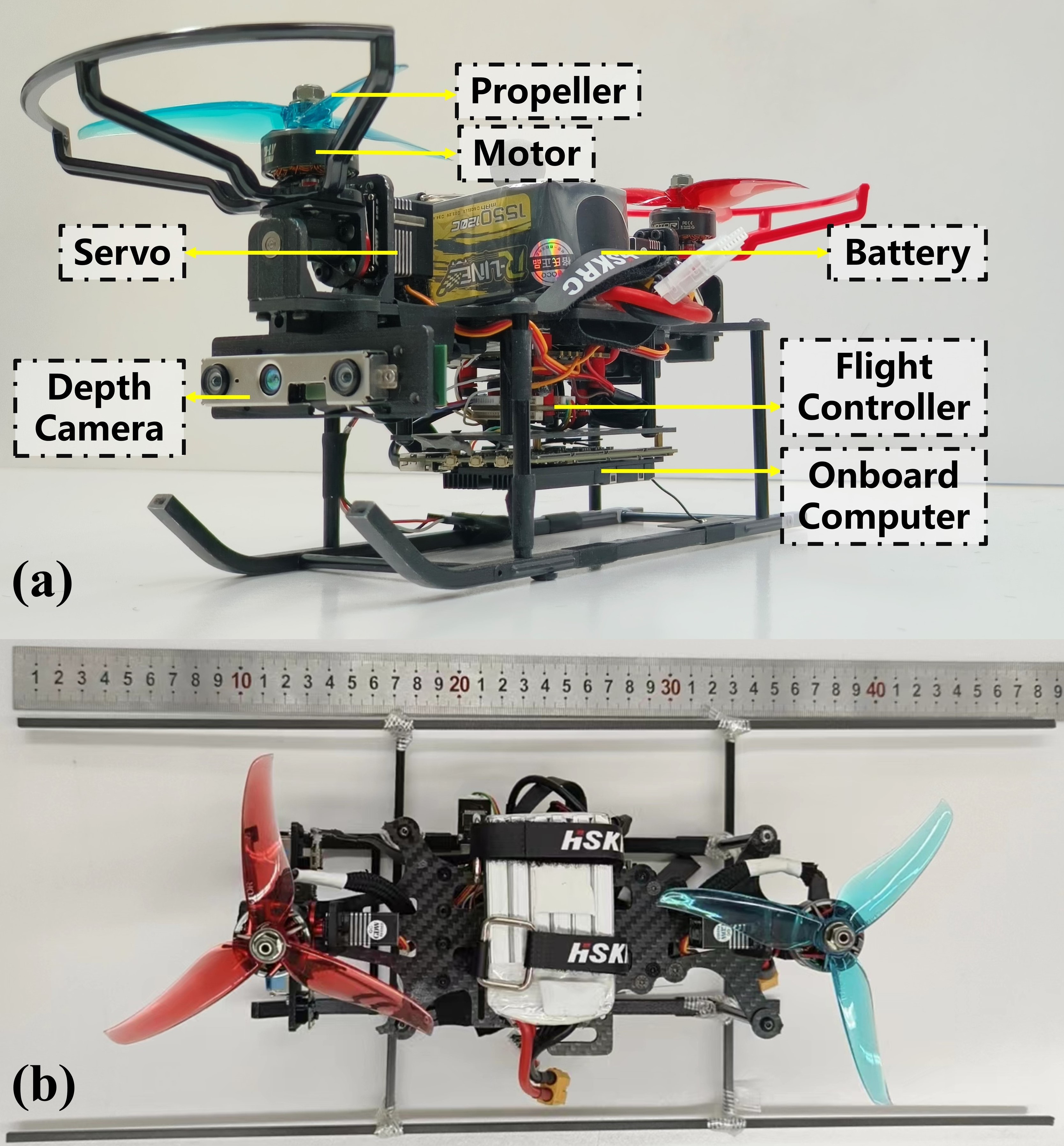}
      \caption{Test platform of proposed bi-copter. (a): System of the bi-copter platform. (b) Bi-copter platform with extended size.}
      \label{fig: testplatform}
\end{figure}

 Bi-copter is known as a type of micro UAV composed of two rotors mounted on a rotation axis with their tilt-angle controlled by two servos. Take the short side as the forward-moving direction gives great traversability due to its elongated shape.  In this work, the proposed test platform is an elongated bi-copter sized $40cm \times 15cm$ as shown in Fig. \ref{fig: testplatform}. The lift and toque were generated by a pair of 5-inch propellers, and the px4 flight controller\cite{px4meier} was adapted to maintain the attitude stabilization. The bi-copter also carried an RK3588 onboard computer for real-time motion planning and trajectory tracking control algorithm deployment.

\subsection{Differential Flatness of Bi-copter UAV}
Differential flatness is a property of using a group of flat variables and their derivatives to express the UAV state $\mathbf{x}$  and its input $\mathbf{u}$. For a differential flat system, there exists a group of variables $\mathbf{y}$, the state $\mathbf{x}$ and input $\mathbf{u}$ can be modeled as the function of the flat variable and its derivatives \cite{levine2011necessary}. For the proposed bi-copter platform, the differential flatness property is similar to \cite{he2022design}. With the assumption of ignoring the servo reaction torque, the system state and the input can be present by the flat variables composed of position and orientation $\mathbf{y}=\left[x,y,z,\psi\right]$ and its derivatives, by planning the position and orientation trajectory, the collision-free motion of bi-copter can be achieved. 

\section{Whole-Body collision free Path Searching}
\label{sec:path}
The kinodynamic path search module inspires our whole-body collision-free path-searching method in \cite{zhou2019fast}. We extended it for orientation planning and whole-body collision check.  For a narrow corridor environment, we set the fly height as constant, according to the differential flatness property, and the planning is performed in $(x,y,\psi) \in \mathbb{SE}(2)$.  As shown in Algorithm \ref{alg:search}, $\mathbf{x}_0$ denote the initial state,  $\mathbf{x}_f$ is the terminal state, $\mathcal{O}$ and $\mathcal{C}$ refer to the open and closed set, respectively. 

\begin{algorithm}
        \textbf{Input:} $\mathbf{x}_0, \mathbf{x}_f$ \;
        
	  Initialize()\;

        \textnormal{\textbf{getInput}}()\;
        
	{
	\While{$\lnot \ \mathcal{O}.\textnormal{\textbf{empty}}()$}{
		$ n_c \gets \mathcal{O}$.\textbf{pop}(), $\mathcal{C} $.\textbf{insert}($ n_c $) \;

		\If{$ \textnormal{\textbf{ReachGoal}}(n_c) \lor \textnormal{\textbf{AnalyticExpand}}(n_c)$}{
			\Return{\textnormal{\textbf{RetrievePath}()}}\;
		}
		$ primitives   \gets $ \textbf{Expand}($ n_c $)\; 
  
		$ nodes \gets \textnormal{\textbf{CheckCollision}}( primitives ) $\;
  
		\For{$ n_i \ \textbf{in} \ nodes $}{
			\If{$ \lnot \ \mathcal{C}.\textnormal{\textbf{contain}}(n_i) \land \textnormal{\textbf{CheckFeasible}}(n_i)  $}{
				$ g_{temp} \gets n_c.g_c + \textnormal{\textbf{EdgeCost}}(n_i)$ \;
    
				\If{$ \lnot \ \mathcal{O}.\textnormal{\textbf{contain}}(n_i) $}{
					$ \mathcal{O} $.\textbf{add}($ n_i $)\;						
				}
				\ElseIf{$ g_{temp} \geq n_i.g_c $}{
						continue\;
				}
				$ n_i.parent \gets n_c, \ n_i.g_c \gets g_{temp} $\;
				
                    $ n_i.f_c \gets n_i.g_c$ + \textbf{Heuristic}($ n_i$)\;
			}
		}
	}
}

\caption{Dynamic feasible whole-body collision free Path Searching }
\label{alg:search}
\end{algorithm}

\subsection{Dynamic Feasible Whole-Body Path Generation}
\label{sec:search}
In Algorithm \ref{alg:search}, the vehicle state is uniquely defined by the position, orientation, and velocity: 
\begin{equation} 
\label{eq:motion_eq}
  \begin{aligned}
    ^\mathcal{W}\mathbf{x}_{k+1}&=\mathbf{A}^\mathcal{W}\mathbf{x}_{k} + \mathbf{B}\mathbf{u}_{k}, \\
    \mathbf{A} = \begin{pmatrix} 
                    1 & 0 & 0 & \tau &0 \\
                    0 & 1 & 0 & 0 & \tau \\
                    0 & 0 & 1 & \tau & 0 \\
                    0 & 0 & 0 & 1 & 0 \\
                    0 & 0 & 0 & 0 & 1 
                  \end{pmatrix}, & 
     \mathbf{B} = \begin{pmatrix} 
                    \frac{1}{2}\tau^2 & 0 & 0 \\
                    0 & \frac{1}{2}\tau^2 & 0  \\
                    0 & 0 & \tau  \\
                    \tau & 0 & 0  \\
                    0 & \tau & 0 
                  \end{pmatrix}.   
  \end{aligned}
\end{equation}
Here, $\tau$ is the sampling time, $^\mathcal{W}\mathbf{x}=\left[x,y,\psi,v_x,v_y\right]^T$ is the position, orientation, and velocity in the world frame. The input $\mathbf{u} = \left[a_x,a_y,\omega_z\right]^T$ is the acceleration and head rotation velocity. Given an initial state $\mathbf{x}(0)$ and control input $\mathbf{u}$, the state $\mathbf{x}(t), t \in \left[0,\tau\right]$ can be obtained. The vehicle  kinematics constraints limit the state and control variables:
\begin{equation}
\label{eq:dynamic_cons}
\left[\begin{array}{c}
-\mathbf{a}_{\max } \\
-\mathbf{v}_{\max } \\
-w_{z \max }
\end{array}\right] \leq\left[\begin{array}{c}
\mathbf{a}(t) \\
\mathbf{v}(t) \\
w_{z}(t)
\end{array}\right] \leq\left[\begin{array}{c}
\mathbf{a}_{\max } \\
\mathbf{v}_{\max } \\
w_{z \max }
\end{array}\right], \quad t \in[0, T] .
\end{equation}

In \textbf{getInput}(), a set of discretized inputs $\mathbf{u}_D \subset \mathcal{U}$ and a set of discretized sampling time $\mathcal{\tau}_D \subset \mathcal{\tau}$ is sampled for node expansion in \textbf{Expand}(). Each input dimension is sampled uniformly in  $[-u_{max},u_{max}]$ 
to get the input units $[-u_{max},-\frac{r-1}{r}u_{max}, \dots,\frac{r-1}{r}u_{max}, u_{max}]$, and the sampling time $[\frac{1}{p}\tau,\dots, \frac{p-1}{p}\tau,\tau]$. Finally, this results in $(2r+1)^3$ input units and $p$ pieces of time units.

In \textbf{Expand}(), given the current state, applying all the input and time primitives, we obtain the $p(2r+1)^3$ trajectory primitives and their terminal node $\mathbf{n}_i$. Then, the whole-body collision checks described in \ref{sec:wbcc} are applied to exclude the primitives that will occur in a collision. Then, the terminal nodes of valid primitives are collected for the following computation. 

In \textbf{EdgeCost()}, like the traditional hybrid A* algorithm, we aim to find the initial path with optimal time and control cost. 
In addition, for a trajectory piece in open space, the heading of bi-copter is not constrained strictly by collision, so the heading of the bi-copter is preferred to be aligned with the tangent direction of the trajectory.
So we define the cost of the trajectory primitives with input $\mathbf{u}$ and duration $\tau$ as $e_c=\left(\left\|\mathbf{u}_d\right\|^2+\rho\right)\tau + \lambda_{t}\left(atan(\frac{v_y}{v_x})-\psi\right)^2$  
So the search cost $g_c$ of the path node $\mathbf{n}_c$ consisting of K primitives is:
\begin{equation}
    \label{eq:g_c}
    g_c= {\textstyle \sum_{k=1}^{K}\left(\left(\left\|\mathbf{u}_{dk}\right\|^2+\rho\right)\tau_k + \lambda_{t}\left(atan(\frac{v_{yk}}{v_{xk}})-\psi_k\right)^2\right)}.
\end{equation}

After obtaining the search cost, we calculate the heuristic function $h_c$ with the same method in \cite{zhou2019fast}. Finally, each node's cost function $f_c$ can be defined as $f_c=g_c+h_c$.

\begin{figure}[t]
     \centering
      \includegraphics[width=.6\linewidth]{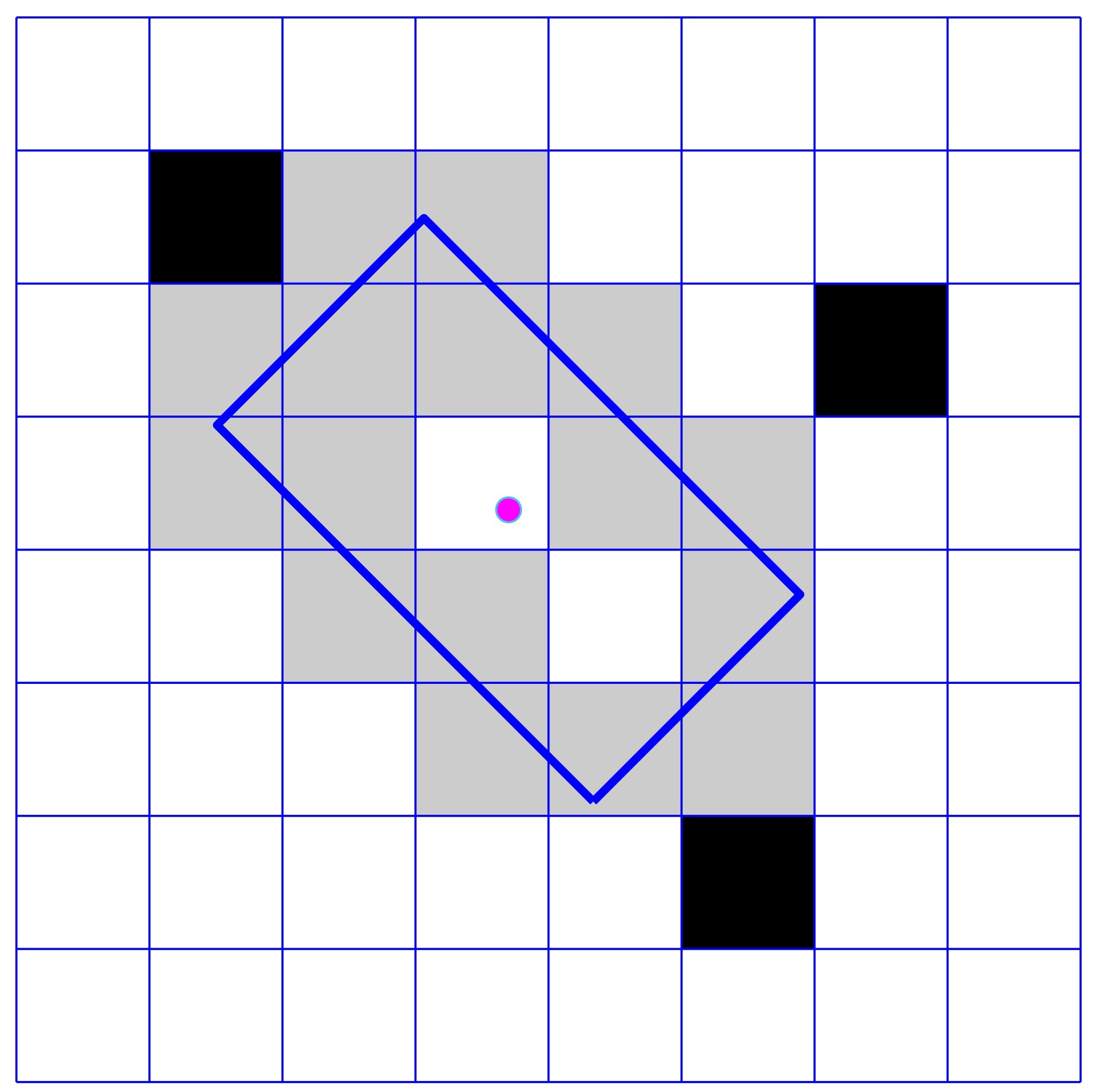}
      \caption{Whole-body collision check of bi-copter, the blue rectangle is the orientated bounding box of the whole-body of bi-copter, black grids are the obstacle, and gray ones are grids that need to be checked for collision.}
      \label{fig: whole_collision}
\end{figure}
\subsection{Whole-Body Collision Check}
\label{sec:wbcc}
As described in Section \ref{sec:search}, during the expansion, the path searching method needs to check if the bi-copter will collide in each path primitive between two path nodes.
For each primitive, given the start state $\mathbf{x}_s$, input $\mathbf{u}$ and duration $\tau$, we will check the collision state of the expand primitive at N discrete times ${\frac{1}{N}\tau},\dots, \tau$, the state at each time is given by \eqref{eq:motion_eq}. After giving the state $\mathbf{x}$ of the bi-copter with known width $l_w$ and length $l_l$, the vertex of the orientated bounding box of the bi-copter can be acquired by \eqref{eq:vertex}.
\begin{equation}
    \label{eq:vertex}
    ^\mathcal{W}\mathbf{v} =\begin{bmatrix}
 cos(\psi) & -sin(\psi )\\
 sin(\psi ) & cos(\psi )
\end{bmatrix} {^\mathcal{B}\mathbf{v}} + \begin{bmatrix}
x \\
y
\end{bmatrix}.
\end{equation}

After obtaining the vertex coordinate in world frame $^\mathcal{W}\mathbf{v}$, the collision can be determined by checking the occupancy status of the grids the box boundary edge laid in, as shown in Fig. \ref{fig: whole_collision}. 
If a grid collision is found in any checkpoint of the primitives, the \textbf{CheckCollision}() of the primitives is returned as false, and the expanded node is marked invalid for the expansion.

\subsection{Safe Flight Corridor Generation}
\label{sec:sfc}
Traditional flight corridor generation methods inflate the free space to generate the polygon corridor based on the initial path, while most methods generate corridors too sparse and for the narrow environment, the intersection of neighbor polygons is not guaranteed to contain full vehicle as shown in Fig. \ref{fig: narrow_com}, which lead to failing of the trajectory optimization. In our method, to benefit the whole-body collision-free path, we first discretize the path and get the sample point as the constrained point used in Section \ref{sec:trajectory}. For each sample point, we have the oriented bounding box of the bi-copter as the initial convex flight corridor. Then, the corridor is generated by expanding and aligning the perpendicular direction of each edge of the bounding box. As a result, the trajectory is guaranteed whole-body collision-free by confining the full vehicle shape or the full edge of the vehicle inside the flight corridor. For each corridor, the H-representation\cite{davidpoly} is used to represent each polygon:
\begin{equation}
    \label{eq:polyhdron}
   \mathcal{P}=\left\{ \mathbf{q}\in \mathbb{R}: \mathbf{A}\mathbf{q} \le \mathbf{b} \right\},
\end{equation}
where $\mathbf{A}=\left(\mathbf{A}_1,\dots,\mathbf{A}_n\right)^T \in \mathbb{R}^{n\times2}$, 
$\mathbf{b} = \left(b_1,\dots,b_n\right)\in \mathbb{R}^{n\times2}$, $\mathbf{A}_i$ refers to the normal vector of each edge of one specific polygon, $b_i$ refers to the point of the edge. For the trajectory at time $t$, the position and the orientation can be specified as $\mathbf{\sigma}$ and $\psi$. Then, the coordinate of the point on the bi-copter's bounding box is given by \eqref{eq:vertex}.
For whole body collision constraint, meaning for each control point, all edge vertices must be inside the flight corridor. Therefore, we have:
\begin{equation}
    \label{eq:wbcf}
    \mathbf{A}^T\left(^\mathcal{W}\mathbf{v}\right)-b, \quad \forall ^\mathcal{B}\mathbf{v} \in \varepsilon,
\end{equation}
here, $\varepsilon$ refers to the vertex set of the edge of the bi-copter bounding box.

\section{Spatial-Temporal Trajectory Planning}
\label{sec:trajectory}
This section presents the spatial-temporal joint optimization formulation for a whole-body trajectory planning algorithm. We formulate the trajectory optimization objective function with whole-body collision-free, smoothness, and dynamic feasibility. To achieve the whole-body collision-free, we sampled points on the edge of the bi-copter-oriented bounding box and formulated the collision penalty function as the sum of the collision states of these sampled checkpoints. As the coordinate of the checkpoint is associated with the position and orientation of the UAVs, a joint optimization on position and orientation is applied to obtain the optimized trajectory. The lightweight quasi-Newton optimizer L-BFGS\cite{liu1989limited} is adopted to solve the numerical optimization.

\subsection{Problem Formulation}
After sampling the trajectory nodes obtained through the search process, we utilize the Minimum Control Effort Polynomial Trajectory, known as MINCO \cite{WANG2022GCOPTER}, as the foundational representation for the trajectories. The trajectory is defined as:
\begin{equation}
  \begin{gathered}
  \mathfrak{T}_{\mathrm{MINCO}}=\left\{p(t):[0, T] \mapsto \mathbb{R}^m \mid \mathbf{c}=\mathcal{M} \mathbf{q}, \mathbf{T}), \right. \\ \left.   \mathbf{q} \in \mathbb{R}^{m(M-1)}, \mathbf{T} \in \mathbb{R}_{>0}^M \right\},
  \end{gathered}
  \label{eq:minco}
\end{equation}
where $\mathbf{q}=\left\{q_1,q_2,\dots,q_{M-1}\right\}$ is the intermediate waypoint set, and $\mathbf{T}=\left\{T_1,T_2,\dots,T_M\right\}$ is the duration of the $i$-th segment.  Define the 3-dimensional segmented polynomial trajectory class $p(t) = [x(t),y(t),\psi(t)]^T \in \mathbb{SE}(2)$  as consisting of $M$ segments of $N=2s-1$ order polynomials. The trajectory for the $i-\textnormal{th}$ segment is defined as:
\begin{equation}
  \label{eq:poly_traj}
  p_i(t)=\mathbf{c}_i^T\beta(t), t\in\left[0,T_i\right],
\end{equation}
where $\mathbf{c} = \left(\mathbf{c}_1^T,\dots,\mathbf{c}_M^T,\right)\in \mathbb{R}^{2Ms \times m}, \mathbf{c}_i\in\ \mathbb{R}^{2s\times m}$ is the coefficient of each segment of the polynomial trajectory.
Given $s=3$, we have a five-order of polynomials, which minimizes the integration of jerk. The objection function of the bi-copter motion planning problem in a narrow corridor environment is formulated as follows:
\begin{subequations}
  \label{eq:obj_func}
  \begin{align}
  \min _{p(t), T} & \mathscr{J}_0 = \int_0^T \left\| x^{\left(3\right)}(t)\right\|^2 d t+\rho T    \label{eq:obj_funca}
  \\
  \text { s.t. }
  & \mathcal{G}_d\left(x(t), \ldots, x^{(s)}(t)\right) \preceq \mathbf{0} \quad \forall t \in[0, T], \label{eq:obj_funcb}\\
  & \mathbf{\varepsilon }(t) \in \mathcal{P} \quad \forall t \in[0, T], \label{eq:obj_funcc}\\
  & p^{[s-1]}(0)=\bar{x}_o, p^{[s-1]}(T)=\bar{x}_f, \label{eq:obj_funcd}
  \end{align}
\end{subequations}
where $\int_0^T \left\| x^{\left(3\right)}(t)\right\|^2 d t$ is the minimum jerk term, which is used to penalize trajectory smoothness, and $\rho T$ is the term to penalize the total time. 
$\mathcal{G}_d$ in \eqref{eq:obj_funcb} is the violation function, indicating the constraint term set. In our formulation the set $\mathcal{D}=\left\{d:d=v,a,w_z\right\}$ includes the whole-body collision avoidance $x$ and dynamic feasibility $\left(v,a,w_z\right)$. 
\eqref{eq:obj_funcc} is the whole-body collision constraint term as described in Section \ref{sec:sfc}.
$\bar{x}_o, \bar{x}_f \in \mathbb{R}^{2 \times s}$ in  \eqref{eq:obj_funcd} is the boundary condition.  

As proved in \cite{WANG2022GCOPTER}, the inequality constraints in (\ref{eq:obj_funcb}) can be formulated by a penalty term $P_{\Sigma}$ and transformed the objection function (\ref{eq:obj_funca}) with inequality constraints to an unconstrained nonlinear optimization problem:
\begin{equation}
  \label{eq:uncons_obj}
   \min _{p(t), T} \mathscr{J} = \int_0^T \left\| x^{\left(3\right)}(t)\right\|^2 d t+\rho T + S_{\Sigma}(\mathbf{c},T).
\end{equation}

To calculate the violation penalty, we discretized each segment of the trajectory by a fixed time-step $\delta T $, the total number of the control point of the given segment with duration $T$ is given by $K=\frac{T}{\delta T}$, then we penalize all the control point violate the constraints in $\mathcal{\mathcal{G}_d}$ to get the penalty term. The whole penalty term of the trajectory with $M$ pieces is:
\begin{equation}
\label{eq:penalty}
\begin{aligned}
& P_{\Sigma}=\sum_{d \in \mathcal{D}} w_d \sum_{i=1}^M \sum_{j=1}^K V_{d, i, j}\left(\mathbf{c}_i, \mathrm{~T}\right), \\
& V_{d, i, j}\left(\mathbf{c}_i, \mathrm{~T}\right)=\delta t \mathrm{~L}_1\left(\mathcal{G}_{d, i, j}\right),
\end{aligned}
\end{equation}
where $w_d$ is the penalty weight of each violation term, and $L_1$ is the L1-norm relaxation function.

\subsection{Optimization Formulation}
To solve the unconstrained nonlinear optimization problem formulated by equation \eqref{eq:uncons_obj}, the gradient of the cost function with variable $\mathbf{c},T$ needs to be modeled, the gradient of the smoothness terms can be acquired as:
\begin{equation}
\begin{aligned}
& \frac{\partial \mathcal{J}_o}{\partial c_i}=2\left(\int_0^{T_i} \beta^{(3)}(t) \beta^{(3)}(t)^{\mathrm{T}} \mathrm{d} t\right) c_i, \\
& \frac{\partial \mathcal{J}_o}{\partial T_i}=c_i^{\mathrm{T}} \beta^{(3)}\left(T_i\right) \beta^{(3)}\left(T_i\right)^{\mathrm{T}} c_i+\rho .
\end{aligned}
\end{equation}

By the chain rule, the gradients of the violation penalize term $S_{\Sigma}$ w.r.t $c_i$ and $T$ are converted to the gradient of $\mathcal{G}_d$ w.r.t $c_i$ and $T$
\subsubsection{Dynamic Feasibility Penalty}
Dynamic feasibility constraints penalty is:
\begin{equation}
\label{eq:dy_feasi_vio}
\begin{cases}\mathcal{G}_w=\left\|w_{zi}(t)\right\|^2-w_{zm}^2 \leq 0, & \forall t \in\left[0, T_i\right], \\ \mathcal{G}_v=\left\|v_i(t)\right\|^2-v_m^2 \leq 0, & \forall t \in\left[0, T_i\right], \\ \mathcal{G}_a=\left\|a_i(t)\right\|^2-a_m^2 \leq 0, & \forall t \in\left[0, T_i\right].\end{cases}
\end{equation}

Then the violation penalty function $V_d$ is
\begin{equation}
\label{eq:dy_feasi_pena}
V_*=\delta tL_1(\max\left[\mathcal{G}_*,0\right]), *=\left\{v,a,w_z\right\}.
\end{equation}

\subsubsection{Whole-body Collision Avoidance Penalty}
For mass-point motion planning methods such as \cite{WANG2022GCOPTER}, the collision avoidance is modeled to constrain the intermediate point inside the flight corridor, according to Section \ref{sec:sfc} $\mathcal{G}_{pm}(\mathbf{p}(t))=\mathbf{A}_i\mathbf{p}(t)-\mathbf{b}_i$.  The whole-body collision avoidance violation is formulated as the sum of the collision avoidance penalty at the sample point on the bounding box edge of the bi-copter.
\begin{equation}
\label{eq:wb_pena}
 \mathcal{G}_p = \sum_{\mathbf{v} \in E}\mathcal{G}_{pm}\left({^\mathcal{W}\mathbf{v}}\right), 
\end{equation}
Here, $\mathbf{v}$ is the sample checkpoint of the bounding box edge $E$ in the body frame of the bi-copter. As can be seen from (\ref{eq:wb_pena}), the collision penalty is associated with the position and orientation of the bi-copter, and the coordinate of the sample checkpoint in the body frame is constant with the given size. So we have the gradient of the collision avoidance penalty violation:
\begin{equation}
\label{eq:wb_pena_gradient}
\begin{aligned}
 \frac{\partial \mathcal{G}_p}{\partial \mathbf{p}} & = \frac{\partial \mathcal{G}_{pm}}{\partial {^\mathcal{W}\mathbf{v}}}
 , \\
 \frac{\partial \mathcal{G}_p}{\partial \psi} & = \frac{\partial \mathcal{G}_{pm}}{\partial {^\mathcal{W}\mathbf{v}}} \frac{\partial{^\mathcal{W}\mathbf{v}}}{\partial \psi} \\ 
  & = \frac{\partial \mathcal{G}_{pm}}{\partial {^\mathcal{W}\mathbf{v}}} \left(\begin{bmatrix}
 -sin(\psi) & -cos(\psi )\\
 cos(\psi ) & -sin(\psi )
\end{bmatrix} {^\mathcal{B}\mathbf{v}} \right) .
\end{aligned}
\end{equation}

\section{Experiments and Results}
\label{sec:experiments}
To validate the efficiency of the proposed method, we performed comparison experiments with other motion planning methods in narrow environments. What's more, a simulation with a different-scale bi-copter was conducted to illustrate the robustness of the proposed method. Finally, real-world experiments were performed in narrow environments with different-size bi-copters. 

\begin{table}[b]
    \centering
    \vspace{0.2cm}
        \caption{\centering Methods Comparison for Non-convex shape robot}
\begin{tabular}{c|c c c}
\hline
Methods & $Optimize Time(s)$  & $Jerk Cost(m^2/s^5)$ & $Length(m)$  \\
\hline
Proposed  & $\textbf{0.26}$ & $\textbf{153.60}$ & $\textbf{19.43}$  \\
\hline
WBFP & $0.87$ & $602.92$ & $19.44$ \\
\hline
\end{tabular}
\vspace{0.1cm}
    \label{tab:table_cmp}
\end{table}

\begin{figure}[h]
     \centering
      \includegraphics[width=.9\linewidth]{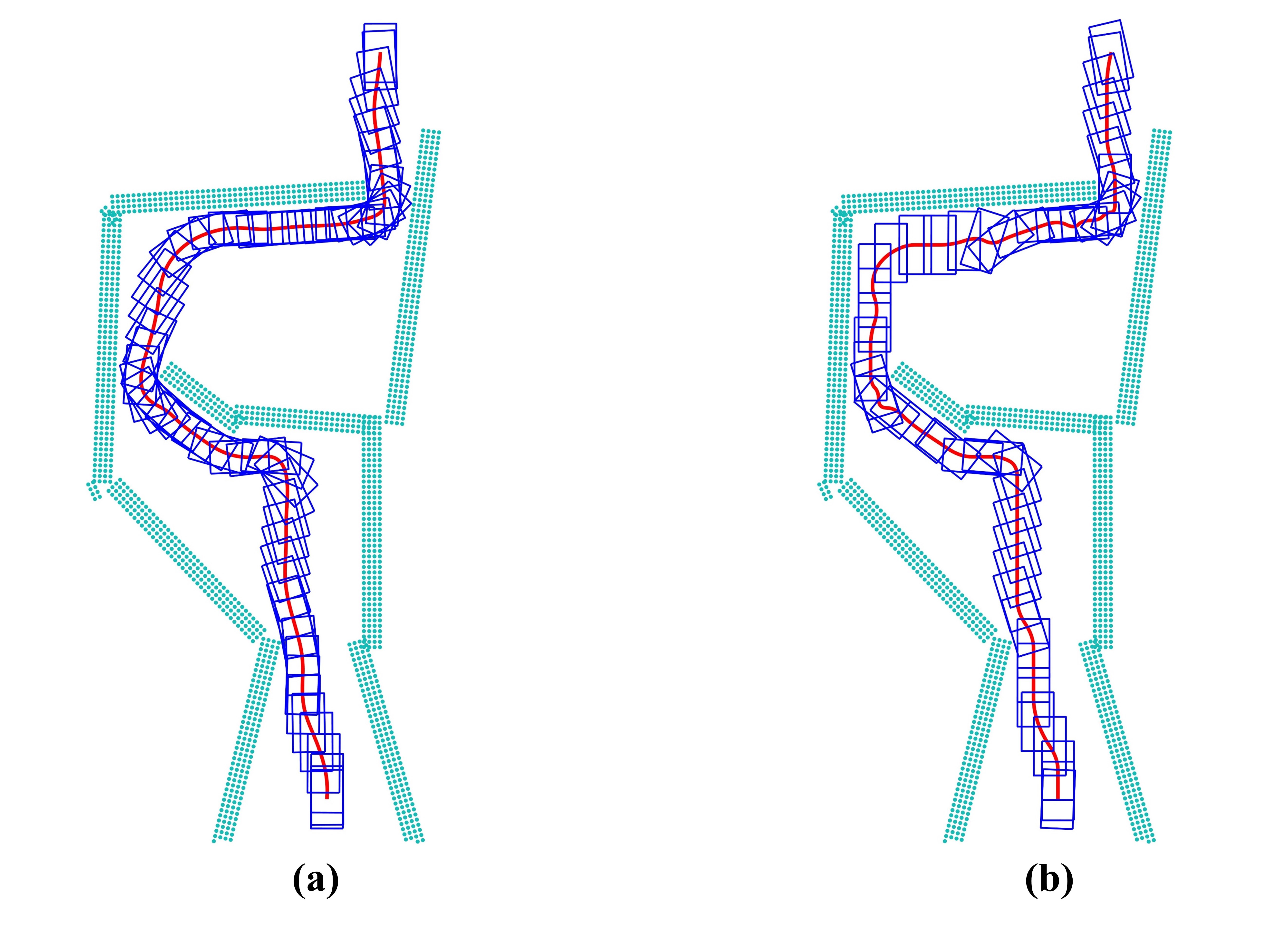}
      \caption{Evaluating the proposed method against WBFP. (a): Proposed method. (b): WBFP }
      \label{fig: bi_wbfp}
\end{figure}

\begin{figure}[h]
     \centering
      \includegraphics[width=\linewidth]{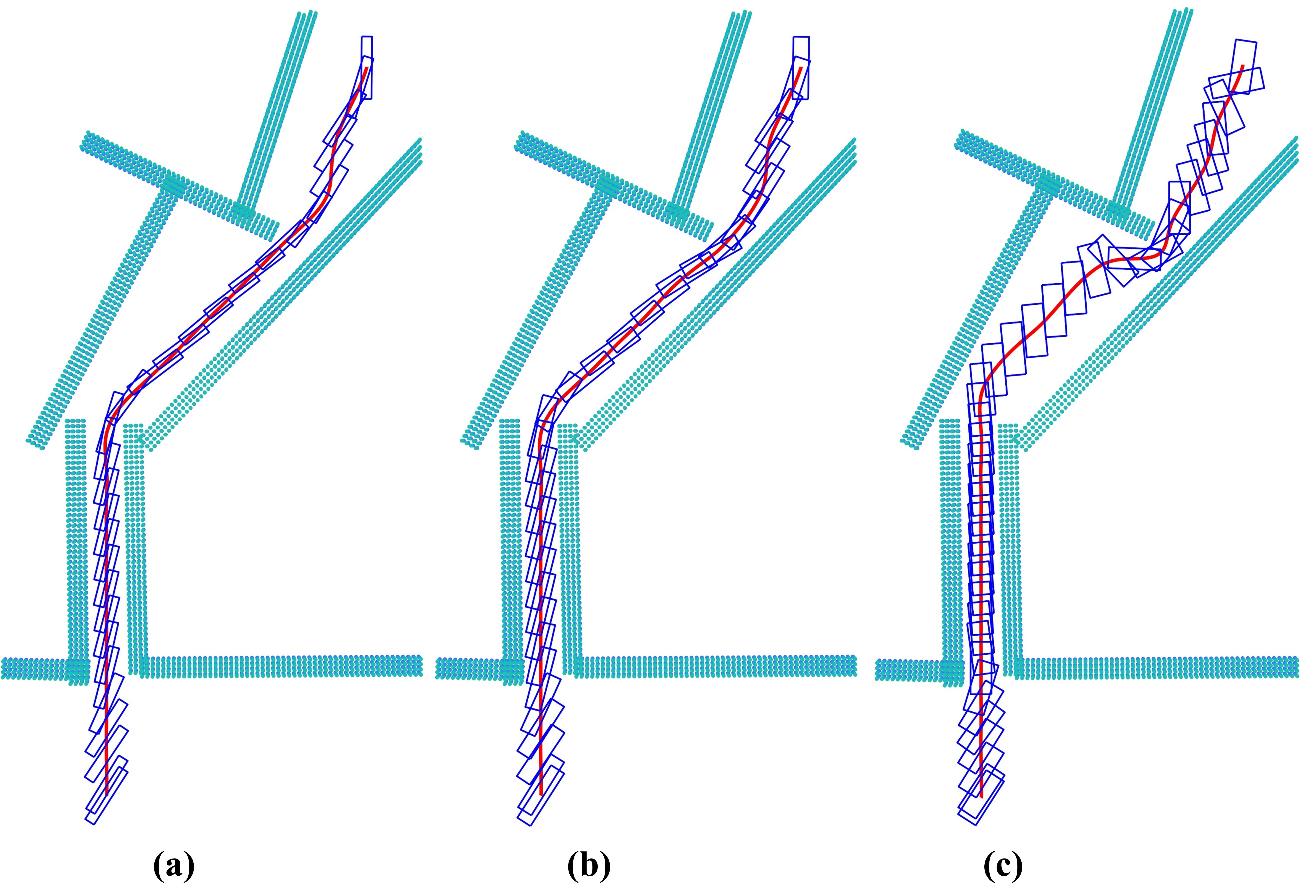}
      \caption{Comparative analysis of varying sizes bi-copters. (a): $0.2m \times 1.2m$. (b): $0.3m \times 1.2m$. (c): $0.4m \times 1.0m$ }
      \label{fig: sim_dif_sz}
\end{figure}

\subsection{Benchmark for Whole-Body Planning}
 Inspired by \cite{geng2023rcesdf}, for whole-body planning condition, we modify the fast-planner \cite{zhou2019fast}, the optimized cost of the collision term was obtained by summing the Euler distance of the sampling point of the whole-body and optimize the position and yaw trajectory jointly, which is called whole-body fast-planner(WBFP). To compare our results more fairly, we implemented the same front-end path generation method and changed the optimizer of the WBFP from NLopt to L-BFGS. We implement both methods in a $8m \times 16m$ environment with the same start and end state. Considering the slimmest site was nearly 0.8m, the size of the robot was set to $0.6m \times 1.2m$. The optimized trajectory was shown in Fig. \ref{fig: bi_wbfp}. As can be seen from the result, both methods can generate collision-free trajectories.  However, the trajectory of the WBFP in such a narrow environment is more curved, while the trajectory by the proposed method has more smoothness. Due to the different trajectory representation forms (polynomial with proposed and B-spline with WBFP). To compare the smoothness fairly, we evaluate the smoothness with the dispersed jerk of the whole trajectory:
 \begin{equation}
\label{eq:jerk_sum}
 J_s=\sum_{i = 0}^{T/\delta t} \left(\left\| p^{(3)}(i\delta t)\right\|^2 + \left\| \psi^{(3)}(i\delta t)\right\|^2 \right)\delta t. 
\end{equation}

The result of the optimized time and jerk cost is shown in Table. \ref{tab:table_cmp}. As can be seen, the trajectory generated by the proposed method has a lower jerk cost. What's more, the optimized time of the WBFP is longer than our method. 

\subsection{Whole-Body Planning with Different Scale}
To compare the robustness of the method, we implement the proposed method in a $15m \times 8m$ environment, and the size of the bi-copter is set to different sizes: $0.2m \times 1.2m$, $0.3m \times 1.2m$ and $0.4m \times 1.0m$, with the narrowest part of the environments is about $0.6m$. All these cases are planned with the same start and end state. The results shown in Fig. \ref{fig: sim_dif_sz} demonstrate that the proposed method can generate a smooth trajectory even when the gap is almost as wide as the bi-copter. As the width of the bi-copter increases, the trajectory becomes more curved, necessitating more rotation to cross the gap.
\begin{figure}[t]
     \centering
      \includegraphics[width=\linewidth]{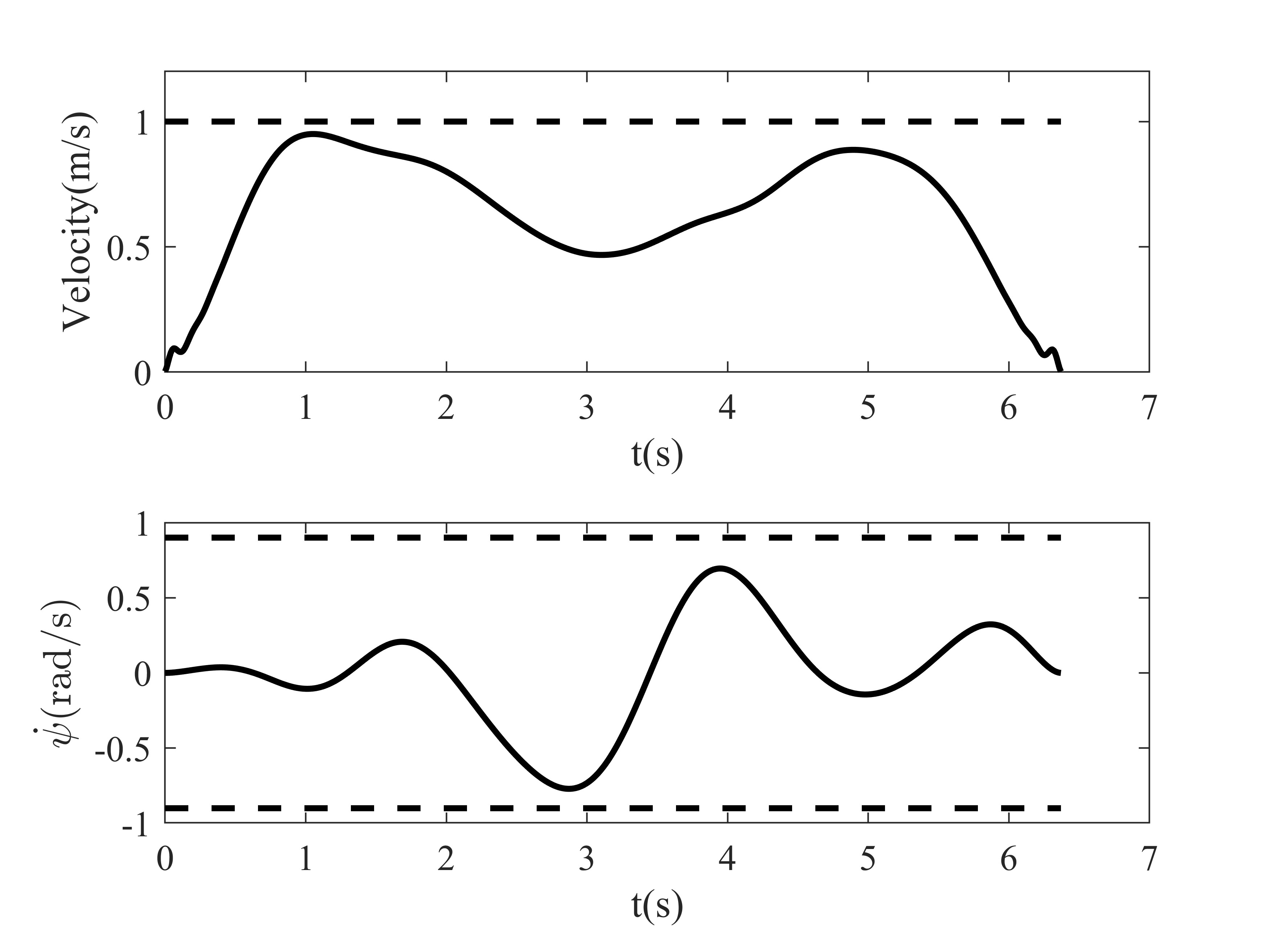}
      \caption{Velocity and yaw velocity profiles of the bi-copter in the extended size experiment case. }
      \label{fig: exp0104_vel}
\end{figure}

\begin{figure*}[!h]
     \centering
      \includegraphics[width=\linewidth]{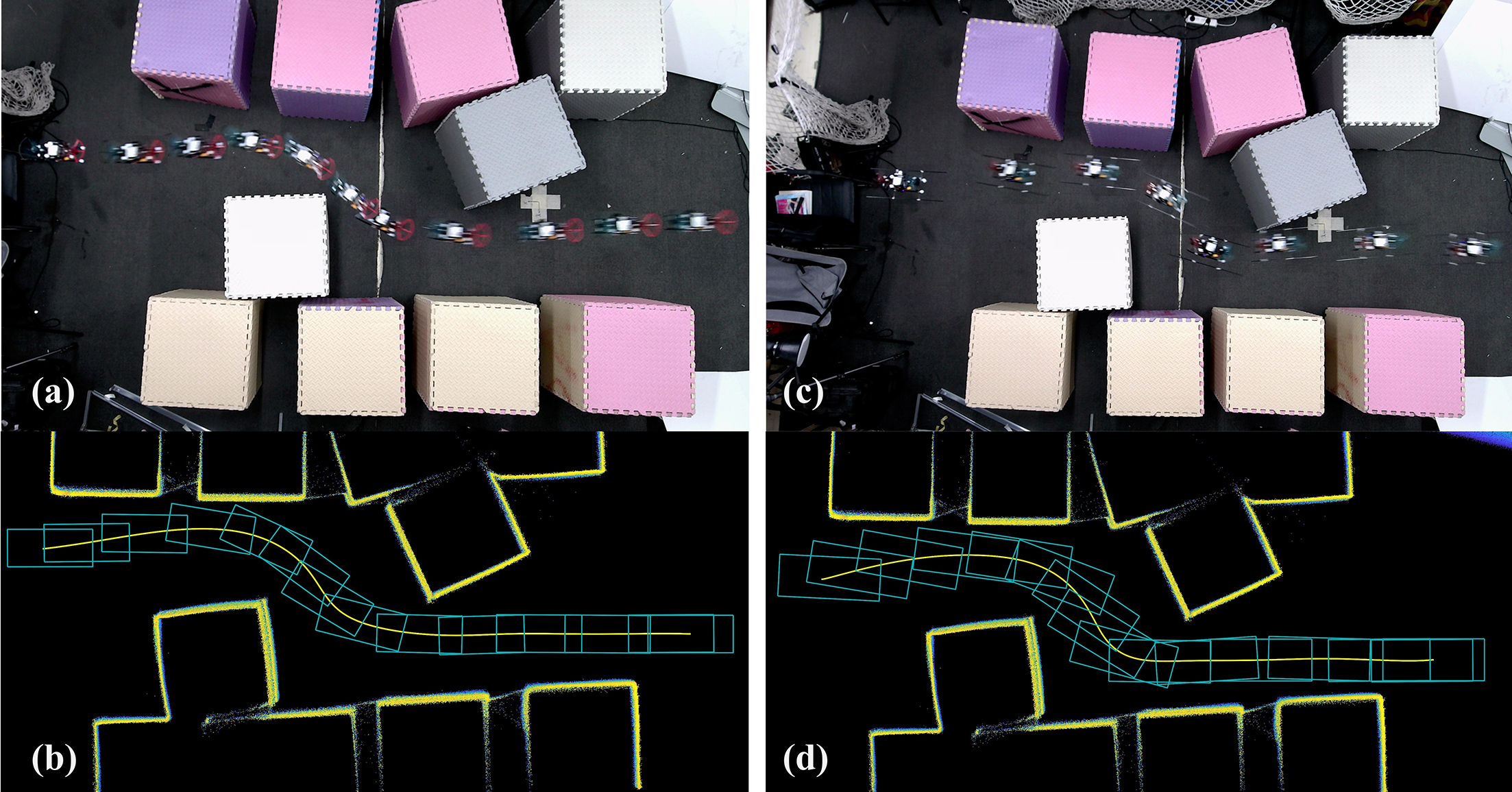}
      \caption{Real-world experiments of the proposed method with different sized bi-copter. (a): Bi-copter with size of $0.15m \times 0.4m$. (b): Bi-copter with size of $0.2m \times 0.5m$.  }
      \label{fig: exp0104}
\end{figure*}

\subsection{Real-world Experiments Setup}
To validate the performance in the real world, we implement the proposed method on a real bi-copter UAV and conduct real-world experiments in several narrow environments. The test platform is shown in Fig. \ref{fig: testplatform}. The size of the test platform is $0.15m \times 0.4m$. To ensure safety, we set the planning width and length in the proposed method as $0.2m$ and $0.5m$. The max velocity, max acceleration, and max yaw velocity are set to $1 m/s, 2m/s^2$ and $1.0 rad/s$. In real-world experiments, the Optitrack motion capture system provided the position estimation, and the point cloud map of the environment was prebuilt and stored in the RK3588 onboard computers. All computations of the planning and tracking control were performed onboard. We implement the trajectory tracking method refer to \cite{Faessler18ral}. To validate the robustness of the proposed method, a bi-copter with extended size was developed for experiments by attaching a framework to the original one. The size of the extended bi-copter is $0.2m \times 0.5m$, and the size in the planning method is set as $0.3m \times 0.6m$ for safety margin. The velocity limit is set to $1m/s$ and the max yaw velocity is set to $0.9 rad/s$.

\subsection{Real-world Experiments Result}
In real-world experiments, the bi-copter successfully generates a trajectory of approximately $4m$ in length to navigate through narrow environments with gaps less than $0.5m$, as demonstrated in Fig. \ref{fig: exp0104}. It is capable of generating a safe trajectory in both cases to cross narrow environments, and it adjusts its orientation to avoid collisions.
Fig. \ref{fig: exp0104_vel} displays the velocity and yaw velocity data that will be analyzed. The proposed method satisfies the velocity and yaw velocity constraints while maintaining high speed and rotation speed.
Real-world experiments demonstrate the performance and robustness of the proposed method for bi-copters flying through narrow environments.

\section{Conclusion}
\label{sec:conclusion}
In this paper, a motion planning method for a rectangle-shaped bi-copter flying through complicated and narrow environments is proposed. Firstly, the proposed method generates a dynamically feasible and collision-free path. Taking this path as the initial reference, the flight corridor is expanded based on the expansion of the whole-body bi-copter-oriented bounding box at the constrained points sampled in the initial path. Then, a method for optimizing collision-free trajectories by jointly optimizing position and orientation is proposed to achieve faster and better trajectories for bi-copter UAVs with a rectangular shape. 
Simulation and real-world experiments were conducted on bi-copters of different sizes in narrow environments to validate the reliability and robustness of the proposed method. 
In the future, we plan to extend the proposed method to more complicated height variant tight spaces in $R(3) \times SO(2)$.

\section*{Acknowledgments}
The author would thank Mr Yiwei Zhang and Hongyi Lu for their great support during this study.
This work is supported by the Academic Excellence Foundation of BUAA
for PhD Students.



\bibliographystyle{IEEEtran}
\bibliography{output}

\begin{thebibliography}{10}
\providecommand{\url}[1]{#1}
\csname url@samestyle\endcsname
\providecommand{\newblock}{\relax}
\providecommand{\bibinfo}[2]{#2}
\providecommand{\BIBentrySTDinterwordspacing}{\spaceskip=0pt\relax}
\providecommand{\BIBentryALTinterwordstretchfactor}{4}
\providecommand{\BIBentryALTinterwordspacing}{\spaceskip=\fontdimen2\font plus
\BIBentryALTinterwordstretchfactor\fontdimen3\font minus \fontdimen4\font\relax}
\providecommand{\BIBforeignlanguage}[2]{{%
\expandafter\ifx\csname l@#1\endcsname\relax
\typeout{** WARNING: IEEEtran.bst: No hyphenation pattern has been}%
\typeout{** loaded for the language `#1'. Using the pattern for}%
\typeout{** the default language instead.}%
\else
\language=\csname l@#1\endcsname
\fi
#2}}
\providecommand{\BIBdecl}{\relax}
\BIBdecl

\bibitem{mahony2012multirotor}
R.~Mahony, V.~Kumar, and P.~Corke, ``Multirotor aerial vehicles: Modeling, estimation, and control of quadrotor,'' \emph{IEEE Robotics and Automation Magazine}, vol.~19, no.~3, pp. 20--32, 2012.

\bibitem{zhou2021fuel}
B.~Zhou, Y.~Zhang, X.~Chen, and S.~Shen, ``Fuel: Fast uav exploration using incremental frontier structure and hierarchical planning,'' \emph{IEEE Robotics and Automation Letters}, vol.~6, no.~2, pp. 779--786, 2021.

\bibitem{feng2023predrecon}
C.~Feng, H.~Li, F.~Gao, B.~Zhou, and S.~Shen, ``Predrecon: A prediction-boosted planning framework for fast and high-quality autonomous aerial reconstruction,'' in \emph{2023 IEEE International Conference on Robotics and Automation (ICRA)}.\hskip 1em plus 0.5em minus 0.4em\relax IEEE, 2023, pp. 1207--1213.

\bibitem{li2023autotrans}
H.~Li, H.~Wang, C.~Feng, F.~Gao, B.~Zhou, and S.~Shen, ``Autotrans: A complete planning and control framework for autonomous uav payload transportation,'' \emph{IEEE Robotics and Automation Letters}, vol.~8, no.~10, pp. 6859--6866, 2023.

\bibitem{Faessler18ral}
M.~Faessler, A.~Franchi, and D.~Scaramuzza, ``Differential flatness of quadrotor dynamics subject to rotor drag for accurate tracking of high-speed trajectories,'' \emph{{IEEE} Robot. Autom. Lett.}, vol.~3, no.~2, pp. 620--626, Apr. 2018.

\bibitem{han2021fastracing}
Z.~Han, Z.~Wang, N.~Pan, Y.~Lin, C.~Xu, and F.~Gao, ``Fast-racing: An open-source strong baseline for $\mathrm{SE}(3)$ planning in autonomous drone racing,'' \emph{IEEE Robotics and Automation Letters}, vol.~6, no.~4, pp. 8631--8638, 2021.

\bibitem{fabris2021geometry}
A.~Fabris, K.~Kleber, D.~Falanga, and D.~Scaramuzza, ``Geometry-aware compensation scheme for morphing drones,'' in \emph{2021 IEEE International Conference on Robotics and Automation (ICRA)}, 2021, pp. 592--598.

\bibitem{qin2020gemini}
Y.~Qin, W.~Xu, A.~Lee, and F.~Zhang, ``Gemini: A compact yet efficient bi-copter uav for indoor applications,'' \emph{IEEE Robotics and Automation Letters}, vol.~5, no.~2, pp. 3213--3220, 2020.

\bibitem{qin2022gemini2}
Y.~Qin, N.~Chen, Y.~Cai, W.~Xu, and F.~Zhang, ``Gemini ii: Design, modeling, and control of a compact yet efficient servoless bi-copter,'' \emph{IEEE/ASME Transactions on Mechatronics}, vol.~27, no.~6, pp. 4304--4315, 2022.

\bibitem{he2022design}
X.~He and Y.~Wang, ``Design and trajectory tracking control of a new bi-copter uav,'' \emph{IEEE Robotics and Automation Letters}, vol.~7, no.~4, pp. 9191--9198, 2022.

\bibitem{zhou2021ego}
X.~Zhou, Z.~Wang, H.~Ye, C.~Xu, and F.~Gao, ``Ego-planner: An esdf-free gradient-based local planner for quadrotors,'' \emph{IEEE Robotics and Automation Letters}, vol.~6, no.~2, pp. 478--485, 2021.

\bibitem{tordesillas2019faster}
J.~Tordesillas, B.~T. Lopez, and J.~P. How, ``Faster: Fast and safe trajectory planner for flights in unknown environments,'' in \emph{2019 IEEE/RSJ International Conference on Intelligent Robots and Systems (IROS)}, 2019, pp. 1934--1940.

\bibitem{penicka2022minimum}
R.~Penicka and D.~Scaramuzza, ``Minimum-time quadrotor waypoint flight in cluttered environments,'' \emph{IEEE Robotics and Automation Letters}, vol.~7, no.~2, pp. 5719--5726, 2022.

\bibitem{zhou2019fast}
B.~Zhou, F.~Gao, L.~Wang, C.~Liu, and S.~Shen, ``Robust and efficient quadrotor trajectory generation for fast autonomous flight,'' \emph{IEEE Robotics and Automation Letters}, vol.~4, no.~4, pp. 3529--3536, 2019.

\bibitem{minisnap2011kumar}
D.~Mellinger and V.~Kumar, ``Minimum snap trajectory generation and control for quadrotors,'' in \emph{2011 IEEE International Conference on Robotics and Automation}, 2011, pp. 2520--2525.

\bibitem{loianno2017estimation}
G.~Loianno, C.~Brunner, G.~McGrath, and V.~Kumar, ``Estimation, control, and planning for aggressive flight with a small quadrotor with a single camera and imu,'' \emph{IEEE Robotics and Automation Letters}, vol.~2, no.~2, pp. 404--411, 2017.

\bibitem{oehler2020whole}
M.~Oehler, S.~Kohlbrecher, and O.~von Stryk, ``Whole-body planning for obstacle traversal with autonomous mobile ground robots,'' in \emph{Advances in Service and Industrial Robotics}, K.~Berns and D.~G{\"o}rges, Eds.\hskip 1em plus 0.5em minus 0.4em\relax Cham: Springer International Publishing, 2020, pp. 250--258.

\bibitem{mittal2022articulated}
M.~Mittal, D.~Hoeller, F.~Farshidian, M.~Hutter, and A.~Garg, ``Articulated object interaction in unknown scenes with whole-body mobile manipulation,'' in \emph{2022 IEEE/RSJ International Conference on Intelligent Robots and Systems (IROS)}, 2022, pp. 1647--1654.

\bibitem{li2022optimization}
B.~Li, T.~Acarman, Y.~Zhang, Y.~Ouyang, C.~Yaman, Q.~Kong, X.~Zhong, and X.~Peng, ``Optimization-based trajectory planning for autonomous parking with irregularly placed obstacles: A lightweight iterative framework,'' \emph{IEEE Transactions on Intelligent Transportation Systems}, vol.~23, no.~8, pp. 11\,970--11\,981, 2022.

\bibitem{ji2022real}
J.~Ji, T.~Yang, C.~Xu, and F.~Gao, ``Real-time trajectory planning for aerial perching,'' in \emph{2022 IEEE/RSJ International Conference on Intelligent Robots and Systems (IROS)}, 2022, pp. 10\,516--10\,522.

\bibitem{han2023efficient}
Z.~Han, Y.~Wu, T.~Li, L.~Zhang, L.~Pei, L.~Xu, C.~Li, C.~Ma, C.~Xu, S.~Shen, and F.~Gao, ``An efficient spatial-temporal trajectory planner for autonomous vehicles in unstructured environments,'' \emph{IEEE Transactions on Intelligent Transportation Systems}, pp. 1--18, 2023.

\bibitem{ma2023decentralized}
C.~Ma, Z.~Han, T.~Zhang, J.~Wang, L.~Xu, C.~Li, C.~Xu, and F.~Gao, ``Decentralized planning for car-like robotic swarm in cluttered environments,'' in \emph{2023 IEEE/RSJ International Conference on Intelligent Robots and Systems (IROS)}, 2023, pp. 9293--9300.

\bibitem{abedini2021robust}
A.~Abedini, A.~A. Bataleblu, and J.~Roshanian, ``Robust backstepping control of position and attitude for a bi-copter drone,'' in \emph{2021 9th RSI International Conference on Robotics and Mechatronics (ICRoM)}, 2021, pp. 425--432.

\bibitem{li2020modeling}
Y.~Li, Y.~Qin, W.~Xu, and F.~Zhang, ``Modeling, identification, and control of non-minimum phase dynamics of bi-copter uavs,'' in \emph{2020 IEEE/ASME International Conference on Advanced Intelligent Mechatronics (AIM)}, 2020, pp. 1249--1255.

\bibitem{albayrak2019design}
Ã.~B. Albayrak, Y.~Ersan, A.~S. Bağbaşı, A.~Turgut~Başaranoğlu, and K.~B. Arıkan, ``Design of a robotic bicopter,'' in \emph{2019 7th International Conference on Control, Mechatronics and Automation (ICCMA)}, 2019, pp. 98--103.

\bibitem{px4meier}
L.~Meier, D.~Honegger, and M.~Pollefeys, ``Px4: A node-based multithreaded open source robotics framework for deeply embedded platforms,'' in \emph{2015 IEEE International Conference on Robotics and Automation (ICRA)}, 2015, pp. 6235--6240.

\bibitem{levine2011necessary}
J.~L{\'e}vine, ``On necessary and sufficient conditions for differential flatness,'' \emph{Applicable Algebra in Engineering, Communication and Computing}, vol.~22, no.~1, pp. 47--90, 2011.

\bibitem{davidpoly}
\BIBentryALTinterwordspacing
D.~AVIS, K.~FUKUDA, and S.~PICOZZI, \emph{ON CANONICAL REPRESENTATIONS OF CONVEX POLYHEDRA}, 2002, pp. 350--360. [Online]. Available: \url{https://www.worldscientific.com/doi/abs/10.1142/9789812777171_0037}
\BIBentrySTDinterwordspacing

\bibitem{liu1989limited}
D.~C. Liu and J.~Nocedal, ``On the limited memory bfgs method for large scale optimization,'' \emph{Mathematical programming}, vol.~45, no. 1-3, pp. 503--528, 1989.

\bibitem{WANG2022GCOPTER}
Z.~Wang, X.~Zhou, C.~Xu, and F.~Gao, ``Geometrically constrained trajectory optimization for multicopters,'' \emph{IEEE Transactions on Robotics}, vol.~38, no.~5, pp. 3259--3278, 2022.

\bibitem{geng2023rcesdf}
S.~Geng, Q.~Wang, L.~Xie, C.~Xu, Y.~Cao, and F.~Gao, ``Robo-centric esdf: A fast and accurate whole-body collision evaluation tool for any-shape robotic planning,'' in \emph{2023 IEEE/RSJ International Conference on Intelligent Robots and Systems (IROS)}, 2023, pp. 290--297.

\end{thebibliography}

\end{document}